\documentclass[11pt]{article}
\usepackage{amsmath,amsfonts,amssymb,amsthm}
\usepackage{graphicx}
\usepackage{cite}
\usepackage{mathtools}
\usepackage{subfig}

\theoremstyle{plain}

\title{Nonlocal Neural Tangent Kernels via Parameter-Space Interactions}
\author{Sriram Nagaraj, Vishakh Hari \\ sriram.nagaraj.atl@gmail.com, vishakh.hari.08@gmail.com} 
\date{\today} 
\date{}

\begin{document}

\maketitle

\begin{abstract}
The Neural Tangent Kernel (NTK) framework has provided deep insights into the training dynamics of neural networks under gradient flow. However, it relies on the assumption that the network is differentiable with respect to its parameters, an assumption that breaks down when considering non-smooth target functions or parameterized models exhibiting non-differentiable behavior. In this work, we propose a Nonlocal Neural Tangent Kernel (NNTK) that replaces the local gradient with a nonlocal interaction-based approximation in parameter space. Nonlocal gradients are known to exist for a wider class of functions than the standard gradient. This allows NTK theory to be extended to nonsmooth functions, stochastic estimators, and broader families of models. We explore both fixed-kernel and attention-based formulations of this nonlocal operator. We illustrate the new formulation with numerical studies.\footnote{This work was independently carried out by the authors in their capacity as independent researchers. The views and content expressed in this article are solely those of the authors and do not reflect those of the Federal Reserve Bank of Cleveland or the Federal Reserve System. }
\end{abstract}

\section{Introduction}

Understanding the training dynamics of neural networks is one of the central goals of theoretical deep learning. The Neural Tangent Kernel (NTK), introduced by Jacot et al. (\cite{jacot}), describes how the output of a neural network evolves during training under gradient flow. Indeed, NTK framework has emerged as a powerful tool for analyzing the training dynamics of wide neural networks. Specifically, the NTK approximates the training dynamics of a neural network by a process of linearizing the network output function $f(x;\theta)$ with input $x$ and with respect to its parameters $\theta$. In the infinite-width limit, this approximation becomes exact.

However, traditional NTK theory assumes infinitesimal parameter updates and linearized training dynamics, which can fall short in capturing nonlocal behavior of real-world networks. Specifically, we list some limitations of the traditional NTK formulation:
\begin{itemize}
    \item It assumes that the function \( f(x; \theta) \) is differentiable with respect to the parameters \( \theta \).
    \item It treats parameter updates via infinitesimal local gradients, which may not capture more global or stochastic behavior.
    \item It fails to handle target functions that are not differentiable (e.g., sign, sawtooth, or other piecewise functions). In practise, however, this issue is handled similar to that of autodifferentiation of non-differentiable functions.
\end{itemize}

In this work, we introduce a nonlocal version of the NTK—called the \textbf{Nonlocal Neural Tangent Kernel (NNTK)}—that generalizes the derivative operator by replacing it with a \emph{nonlocal gradient} in parameter space. This operator integrates finite-difference information across neighborhoods of the parameter vector, weighted by an interaction kernel. Importantly, it is well-defined even when \( f \) is not differentiable in \( \theta \).

We further propose a stochastic approximation of the Nonlocal NTK using Monte Carlo sampling over random parameter perturbations. To increase flexibility and adaptivity, we introduce an attention-based mechanism that assigns learned weights to each sampled perturbation. This leads to our main contribution: an attention-weighted Monte Carlo estimator of the Nonlocal NTK, with learnable number of samples $N$.

\section{Background and Motivation}
Given a neural network $f(x; \theta)$ with parameters $\theta \in \mathbb{R}^d$, the classical NTK is defined as:
\[
K(x, x') = \nabla_\theta f(x; \theta)^\top \nabla_\theta f(x'; \theta).
\]
This form of the NTK is obtained from gradient flow of the underlying function. Indeed, since \( f(x; \theta) \) is a neural network parameterized by \( \theta \), with \( L(\theta) \) being a loss function, under gradient descent, the parameters evolve as:
\[
\frac{d\theta}{dt} = -\nabla_\theta L(\theta).
\]
The evolution of the output at a point \( x \) is given by the chain rule:
\[
\frac{df(x; \theta)}{dt} = \nabla_\theta f(x; \theta)^\top \frac{d\theta}{dt} = -\nabla_\theta f(x; \theta)^\top \nabla_\theta L(\theta).
\]
In supervised learning, \( L(\theta) = \sum_{i=1}^n \ell(f(x_i; \theta), y_i) \), and hence, this yields \[
\frac{d}{dt} f(x; \theta_t) = -\sum_{x'} K(x, x') \frac{\partial \ell}{\partial f(x')}.
\]

\subsection{Need for Nonlocal Version}
The infinitesimal gradient assumption fails to account for finite and nonlocal parameter changes, and also assumes that both $L(\cdot)$ and $f(\cdot)$ are differentiable. In many practical scenarios, this is not true, and we must resort to nondifferentiable loss functions or network outputs. Second, even if the loss/network functions are differentiable, we may desire the inclusion of neighborhood values in a nonlocal fashion for averaging/smoothing purposes. This would necessitate the use of nonlocal version of the gradient operators to develop a NTK-like object in the nonlocal regime. 

\section{Nonlocal Parameter Derivatives}

To further this discussion, we define a \textbf{nonlocal gradient operator} over parameter space. For a function \( f(x; \theta) \), define:
\[
\mathcal{G}_\gamma f(x; \theta) := \int_{\mathbb{R}^d}   \frac{(f(x; \theta') - f(x; \theta))(\theta' - \theta)}{\|\theta' - \theta\|^2} \rho_\gamma(\theta, \theta')\, d\theta',
\]
where \( \rho_\gamma \) is a symmetric, positive kernel that concentrates around \( \theta \) as \( \gamma \to 0 \) (e.g., a Gaussian kernel):
\[
\rho_\gamma(\theta, \theta') = \frac{1}{Z} \exp\left( -\frac{\|\theta - \theta'\|^2}{\gamma^2} \right).
\]
This nonlocal operator is well-defined even when \( f \) is not differentiable in \( \theta \), as it uses finite differences over a neighborhood. We refer to \cite{du} and \cite{mengesha} for the mathematical exposition of nonlocal operators used here.

It can be interpreted as the conditional expectation of a \textbf{stochastic  derivative}, where we sample perturbed parameters \( \theta' \sim \rho_\gamma \) and compute an average change in \( f \):

\[
\mathcal{G}_\gamma f(x; \theta) = \mathbb{E}_{\Theta'\sim \rho_\gamma}[\frac{(f(x; \Theta') - f(x; \theta))(\Theta' - \theta)}{\|\Theta' - \theta\|}] \,
\]
 In the limit \( \gamma \to 0 \), this operator converges (in a weak sense) to the classical derivative, under suitable regularity (see \cite{mengesha, du}).
 Note that a directional version of the above formulation may also be specified where both the high dimensional integration in $\mathbb{R}^d$ as well as the symmetric kernel requirement may be dropped (see \cite{nagaraj24}):
\[
\mathcal{D}_{\delta,\phi} f(\theta) := \int \frac{f(\theta'+t\phi) - f(\theta)}{t} \rho_\delta(t) dt.
\]
This represents a nonlocal directional derivative at $\theta$ in the direction of $\phi$, and is a single dimensional nonlocal operator. We will consider the full nonlocal gradient $\mathcal{G}_\gamma f(x; \theta)$ in subsequent discussions, though the directional alternative can be easily substituted in its place.
\subsection{Nonlocal Gradient Perspective}
Let $\rho_\delta(\theta, \theta')$ be the kernel defining an interaction neighborhood in parameter space. Recall that the nonlocal gradient of $f$ at $\theta$ can be expressed as:
\[
\mathcal{G}_\gamma f(x; \theta) = \int_{\mathbb{R}^d}   \frac{(f(x; \theta') - f(x; \theta))(\theta' - \theta)}{\|\theta' - \theta\|^2} \rho_\gamma(\theta, \theta')\, d\theta'
\]
We define the Nonlocal Neural Tangent Kernel as:
\[
K_\gamma(x, x') := \mathcal{G}_\gamma f(x; \theta)^\top \mathcal{G}_\gamma f(x'; \theta).
\]

This can be computed either analytically (when possible) or estimated using Monte Carlo sampling of \( \theta' \) from the kernel \( \rho_\gamma \).

Compared to the classical NTK, this kernel captures a broader spectrum of functional changes and is capable of operating in nondifferentiable or noisy settings. It generalizes the local gradient structure to a weighted average of directional perturbations.
Reformulating the nonlocal gradient as an expected value (as we saw earlier) motivates defining another form of the Nonlocal NTK:
\[
K_\gamma(x, x') := \mathbb{E}_{\Delta \sim \rho_\gamma} \left[ \frac{f(x; \theta + \Delta) - f(x; \theta)}{\|\Delta\|} \cdot \frac{f(x'; \theta + \Delta) - f(x'; \theta)}{\|\Delta\|} \right].
\]
Here, we assume a sacalar valued function $f(x,\theta)$ and  $\theta$ is fixed within the expectation. Note that instead of a factor $\frac{\Delta}{\|\Delta\|^2}$, we use $\frac{1}{\|\Delta\|}$ in each factor of $K_\gamma(x, x')$. This has the same scaling and asymptotic features, and indeed, had we defined \[
K_\gamma(x, x') := \mathbb{E}_{\Delta \sim \rho_\gamma} \left[ \frac{(f(x; \theta + \Delta) - f(x; \theta))\Delta^\top}{\|\Delta\|^2} \cdot \frac{(f(x'; \theta + \Delta) - f(x'; \theta))\Delta}{\|\Delta\|^2} \right], 
\]we would have obtained the same expression.
\subsection{Monte Carlo Approximation}
We can go one step further, and approximate the Nonlocal NTK using $N$ random perturbations $\{\Delta_k\}_{k=1}^N \sim \rho$:
\[
\hat{K}(x, x') = \frac{1}{N} \sum_{k=1}^N \frac{\delta_k(x) \cdot \delta_k(x')}{\|\Delta_k\|^2},
\]
where $\delta_k(x) := f(x; \theta + \Delta_k) - f(x; \theta)$. This estimator can be computed without access to gradients, making it amenable to black-box settings, in the sense of zeroth-order methods.

\subsection{Attention-Based Weighting}

Instead of a fixed kernel \( \rho_\gamma \), we can use an \textbf{attention mechanism} to adaptively weight the importance of each \( \theta' \). For example, for a fixed collection of $\theta_j$ acting as ``anchors", we can define:
\[
\rho_\gamma^{\text{att}}(\theta, \theta') = \frac{\exp\left( -\frac{\|f(x; \theta) - f(x; \theta')\|^2}{\gamma^2} \right)}{\sum_j \exp\left( -\frac{\|f(x; \theta) - f(x; \theta_j)\|^2}{\gamma^2} \right)},
\]
which gives higher weight to parameter perturbations that result in similar outputs. This encourages the model to attend to directions that preserve output geometry.

\subsection{Nonlocal Gradient Flow Dynamics}
Under nonlocal dynamics, the parameter update becomes:
\[
\frac{d\theta}{dt} = -\mathcal{G}_\gamma L(\theta),
\]
which is the nonlocal analog of gradient flow. When \( L \) is not differentiable, the operator \( \mathcal{G}_\gamma L \) still exists under mild assumptions.

This formulation can also be interpreted as a smoothed version of training dynamics that respects a neighborhood in parameter space, leading to potentially more stable training and better generalization.

\section{Experiments}
We illustrate the foregoing discussion with numerical examples. We first compare the heatmap of the various NTKs: standard, attention based and Monte Carlo with perturbations drawn from uniform/Gaussian distributions. We use challenging nonsmooth target functions $f(x)$ such as:
\begin{itemize}
    \item \( f(x) = \text{sign}(x) \)
    \item \( f(x) = x - \lfloor x \rfloor \) (sawtooth)
\end{itemize} These functions are not differentiable, so a standard NTK would struggle to capture their behavior. We then show how the spectrum of the NTK i.e., the spectrum of the Kernel matrices \( K(x_i, x_j) \) evolves. Finally, we consider the kernel alignment:
\[
\text{alignment}(K, y) = \frac{y^\top K y}{\|K\|_F \cdot \|y\|^2}.
\]and plot this for both the standard and nonlocal NTK in the case of the sign function. All these numerical examples are provided in the appendix.
\section{Discussion}
We introduced a framework for extending NTK to nonsmooth regimes via nonlocal derivatives in parameter space. This leads to a generalized NTK (NNTK) that is both theoretically motivated and empirically powerful. We illustrated our ideas using numerical examples of nonsmooth network functions, and compared the spectrum of NTK matrices in the standard vs. nonlocal regimes.

Future work includes:
\begin{itemize}
    \item Theoretical convergence guarantees for NNTK.
    \item Extension to convolutional networks and Transformers.
    \item Applications to adversarial robustness and Bayesian models.
\end{itemize}

\bibliographystyle{plain}

\begin{thebibliography}{9}
\bibitem{jacot}
A. Jacot, F. Gabriel, and C. Hongler. Neural Tangent Kernel: Convergence and Generalization in Neural Networks. In NeurIPS, 2018.

\bibitem{mengesha}
T. Mengesha and D. Spector. Nonlocal vector calculus and nonlocal balance laws, Journal of Mathematical Analysis and Applications, 2014.

\bibitem{du}
Q. Du, M. Gunzburger, R. Lehoucq, and K. Zhou. A nonlocal vector calculus, nonlocal volume-constrained problems, and nonlocal balance laws, Mathematical Models and Methods in Applied Sciences, 2012.

\bibitem{nagaraj24}
S Nagaraj and T. Hickok, BrowNNe: Brownian Nonlocal Neurons \& Activation Functions, https://arxiv.org/abs/2406.15617
\end{thebibliography}

\section{Appendix}

\begin{figure}%
    \centering
    \subfloat[\centering]{{\includegraphics[width=10cm]{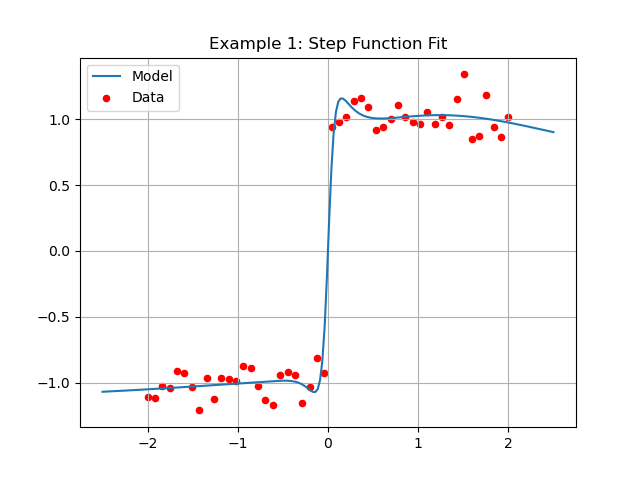} }}%
    \qquad
    \subfloat[\centering]{{\includegraphics[width=5cm]{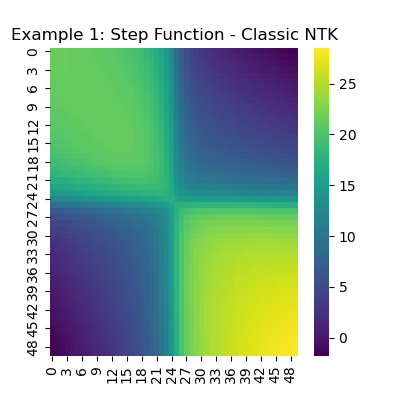} }}%
    \caption{Step Function}%
    \label{fig:example}%
\end{figure}
\begin{figure}%
    \centering
    \subfloat[\centering]{{\includegraphics[width=5cm]{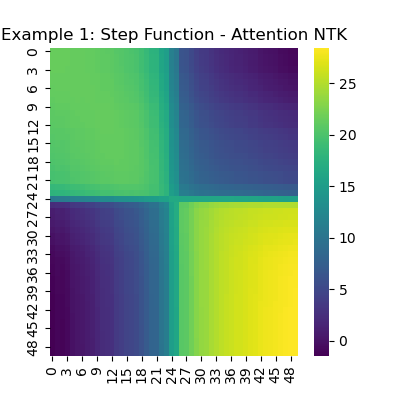} }}%
    \qquad
    \subfloat[\centering]{{\includegraphics[width=5cm]{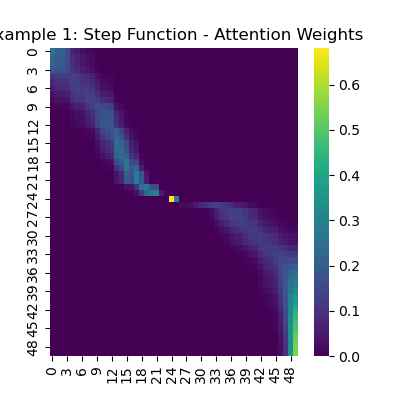} }}%
    \caption{Step Function: Attention}%
    \label{fig:example}%
\end{figure}
\begin{figure}%
    \centering
    \subfloat[\centering]{{\includegraphics[width=5cm]{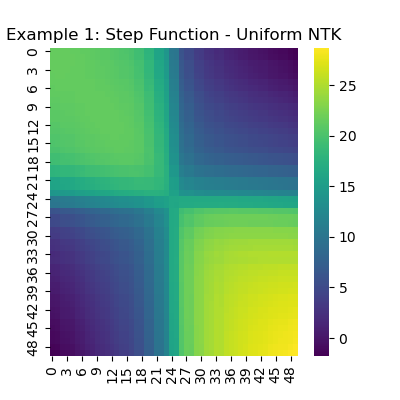} }}%
    \qquad
    \subfloat[\centering]{{\includegraphics[width=5cm]{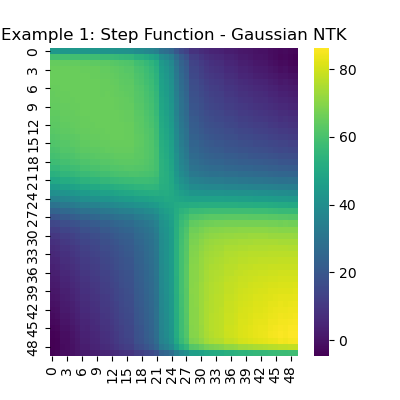} }}%
    \caption{Step Function: Monte Carlo}%
    \label{fig:example}%
\end{figure}

\begin{figure}%
    \centering
    \subfloat[\centering]{{\includegraphics[width=10cm]{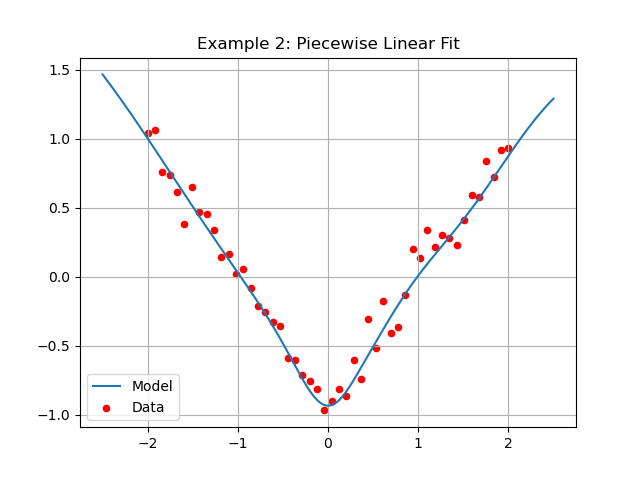} }}%
    \qquad
    \subfloat[\centering]{{\includegraphics[width=5cm]{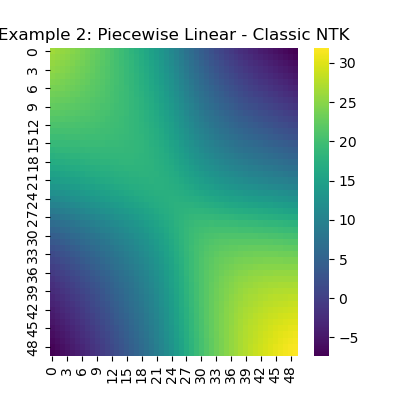} }}%
    \caption{Piecewise Function}%
    \label{fig:example}%
\end{figure}
\begin{figure}%
    \centering
    \subfloat[\centering]{{\includegraphics[width=5cm]{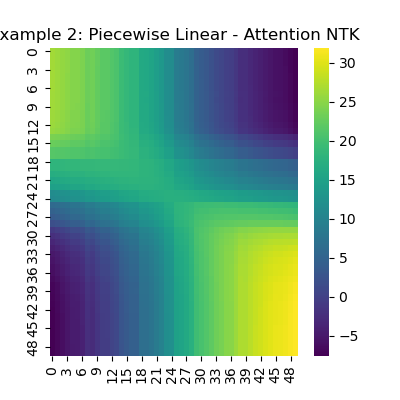} }}%
    \qquad
    \subfloat[\centering]{{\includegraphics[width=5cm]{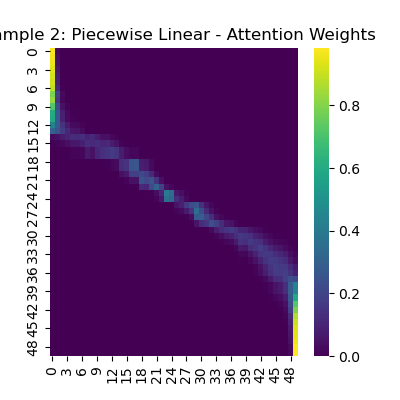} }}%
    \caption{Piecewise Function: Attention}%
    \label{fig:example}%
\end{figure}
\begin{figure}%
    \centering
    \subfloat[\centering]{{\includegraphics[width=5cm]{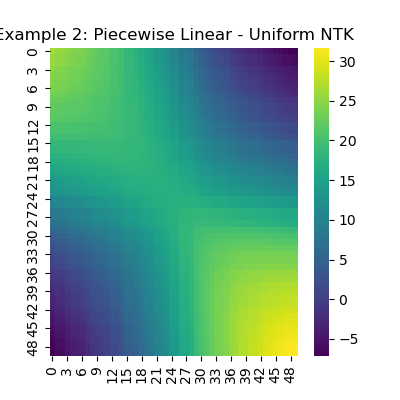} }}%
    \qquad
    \subfloat[\centering]{{\includegraphics[width=5cm]{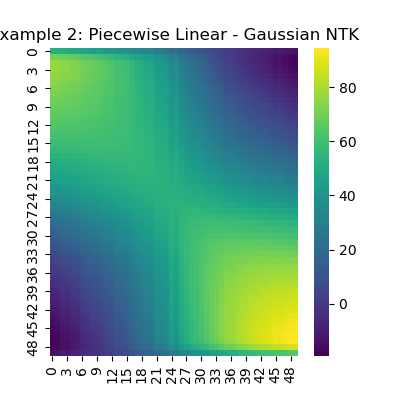} }}%
    \caption{Piecewise Function: Monte Carlo}%
    \label{fig:example}%
\end{figure}

\begin{figure}%
    \centering
    \subfloat[\centering]{{\includegraphics[width=10cm]{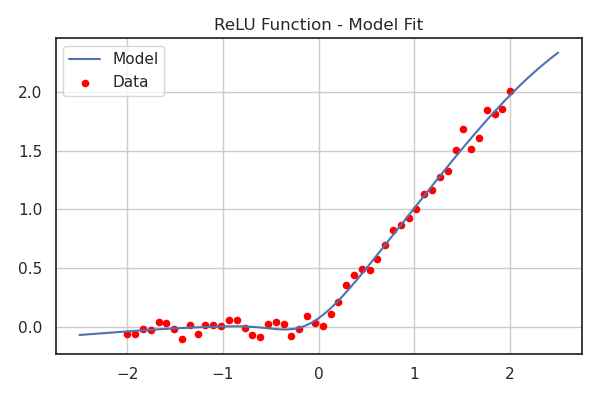} }}%
    \qquad
    \subfloat[\centering]{{\includegraphics[width=5cm]{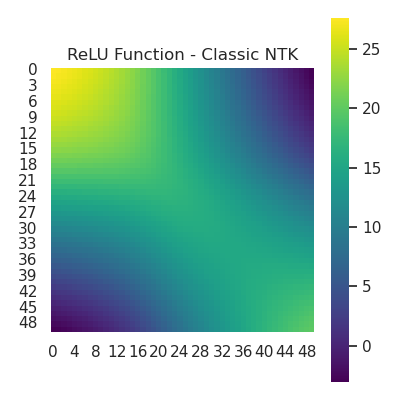} }}%
    \caption{ReLU Function}%
    \label{fig:example}%
\end{figure}
\begin{figure}%
    \centering
    \subfloat[\centering]{{\includegraphics[width=5cm]{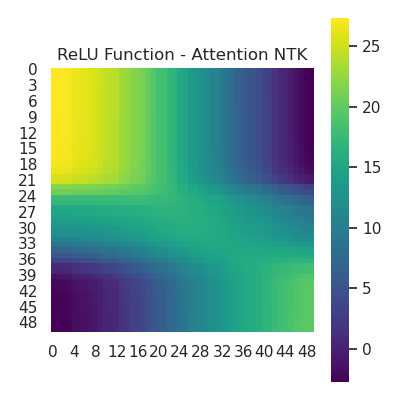} }}%
    \qquad
    \subfloat[\centering]{{\includegraphics[width=5cm]{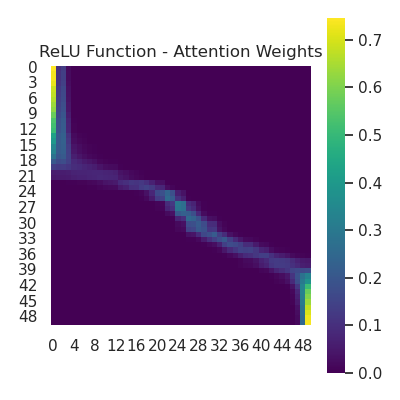} }}%
    \caption{ReLU Function: Attention}%
    \label{fig:example}%
\end{figure}

\begin{figure}%
    \centering
    \subfloat[\centering]{{\includegraphics[width=10cm]{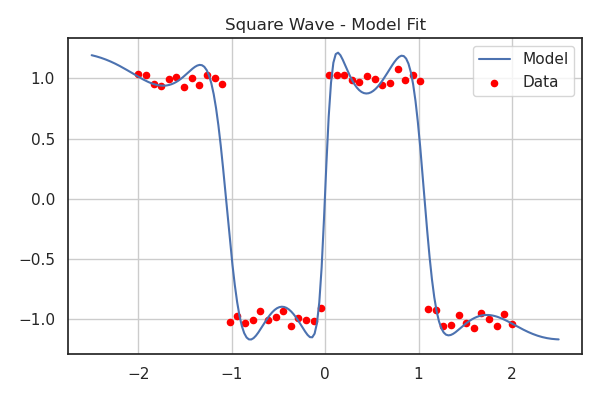} }}%
    \qquad
    \subfloat[\centering]{{\includegraphics[width=5cm]{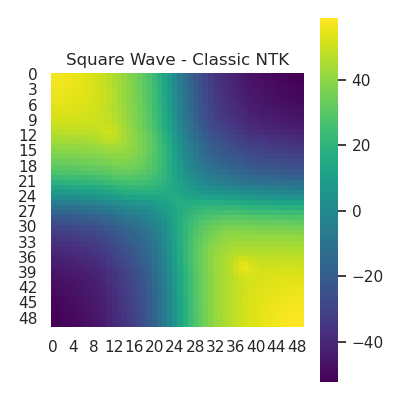} }}%
    \caption{Square Wave Function}%
    \label{fig:example}%
\end{figure}
\begin{figure}%
    \centering
    \subfloat[\centering]{{\includegraphics[width=5cm]{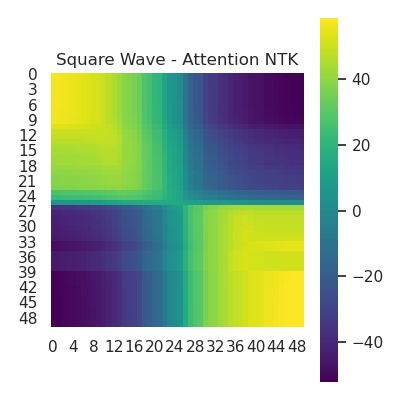} }}%
    \qquad
    \subfloat[\centering]{{\includegraphics[width=5cm]{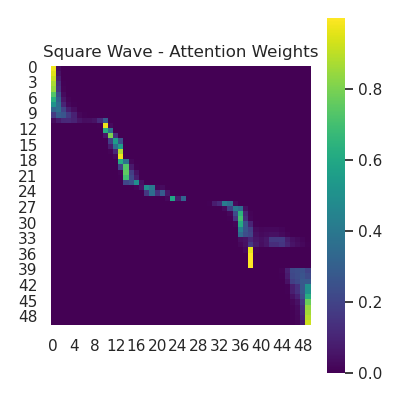} }}%
    \caption{Square Wave Function: Attention}%
    \label{fig:example}%
\end{figure}

\begin{figure}%
    \centering
    \subfloat[\centering]{{\includegraphics[width=10cm]{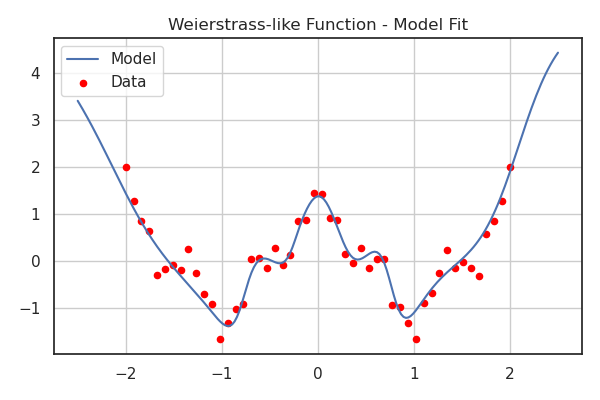} }}%
    \qquad
    \subfloat[\centering]{{\includegraphics[width=5cm]{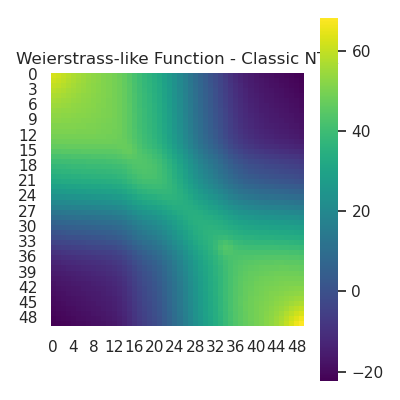} }}%
    \caption{Weierstrass-like Function}%
    \label{fig:example}%
\end{figure}
\begin{figure}%
    \centering
    \subfloat[\centering]{{\includegraphics[width=5cm]{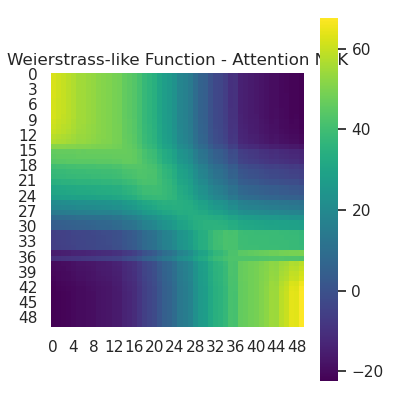} }}%
    \qquad
    \subfloat[\centering]{{\includegraphics[width=5cm]{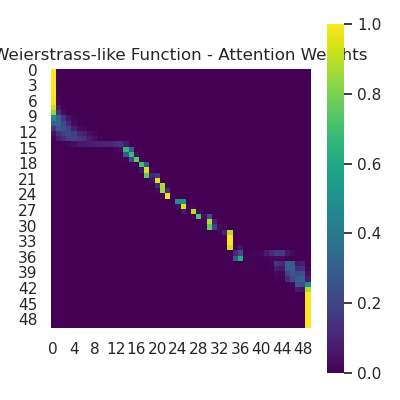} }}%
    \caption{Weierstrass-like Function: Attention}%
    \label{fig:example}%
\end{figure}

\begin{figure}%
    \centering
    \subfloat[\centering]{{\includegraphics[width=10cm]{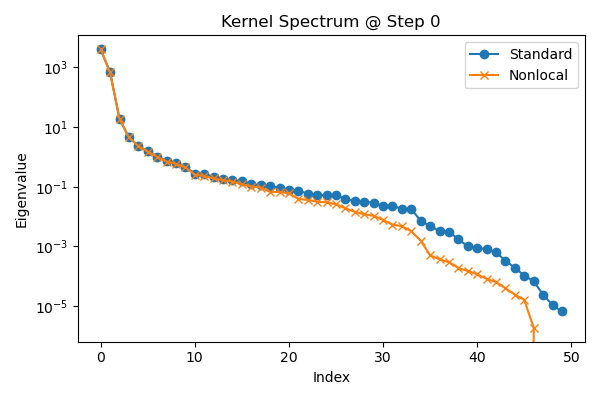} }}%
    \qquad
    \subfloat[\centering]{{\includegraphics[width=10cm]{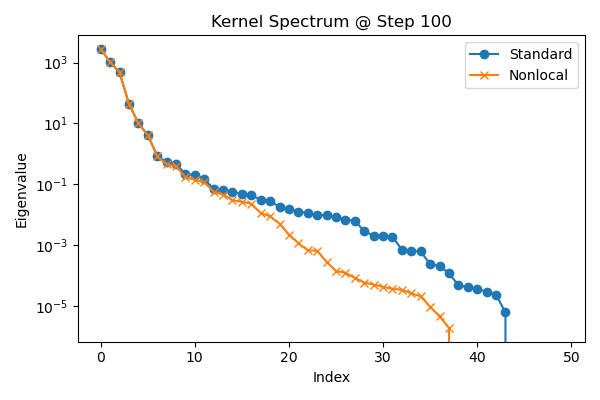} }}%
    \caption{Spectrum Convergence}%
    \label{fig:example}%
\end{figure}

\begin{figure}%
    \centering
    \subfloat[\centering]{{\includegraphics[width=10cm]{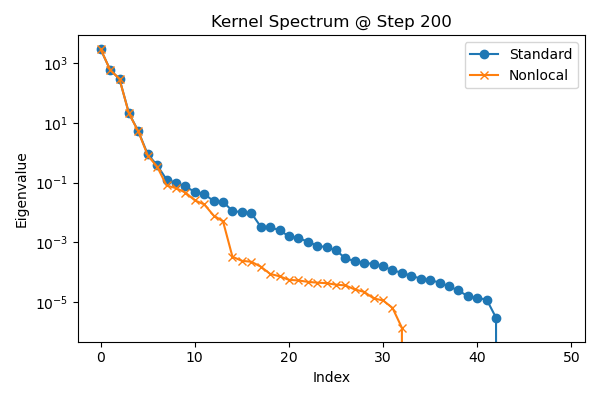} }}%
    \qquad
    \subfloat[\centering]{{\includegraphics[width=10cm]{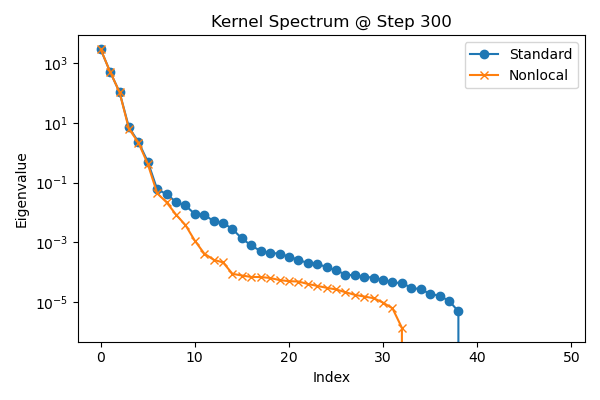} }}%
    \caption{Spectrum Convergence}%
    \label{fig:example}%
\end{figure}

\begin{figure}%
    \centering
    \subfloat[\centering]{{\includegraphics[width=10cm]{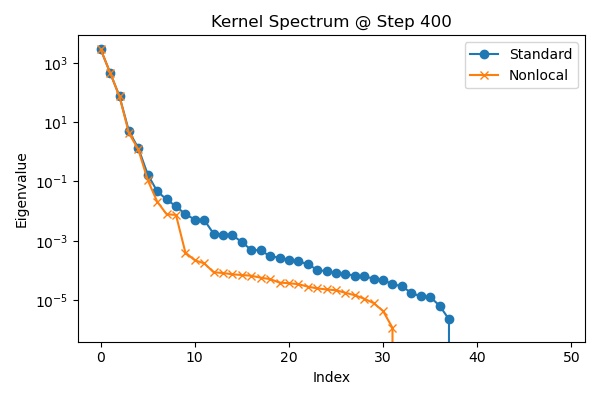} }}%
    \qquad
    \subfloat[\centering]{{\includegraphics[width=10cm]{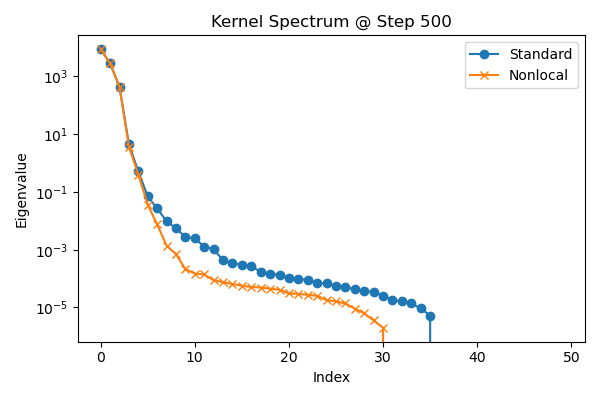} }}%
    \caption{Spectrum Convergence}%
    \label{fig:example}%
\end{figure}

\begin{figure}%
    \centering
    \subfloat[\centering]{{\includegraphics[width=10cm]{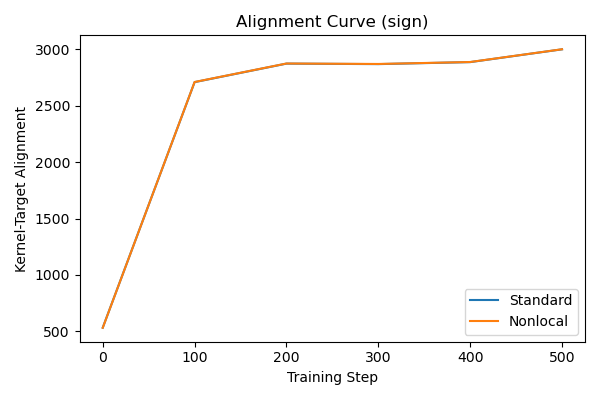} }}%
    \caption{Alignment Curves}%
    \label{fig:example}%
\end{figure}

\end{document}